# Multilevel Image Thresholding Using a Fully Informed Cuckoo Search Algorithm

Xiaotao Huang, Liang Shen*, Chongyi Fan, Jiahua zhu and Sixian Chen

*Abstract*—**Though effective in the segmentation, conventional multilevel thresholding methods are computationally expensive as exhaustive search are used for optimal thresholds to optimize the objective functions. To overcome this problem, population-based metaheuristic algorithms are widely used to improve the searching capacity. In this paper, we improve a popular metaheuristic called cuckoo search using a ring topology based fully informed strategy. In this strategy, each individual in the population learns from its neighborhoods to improve the cooperation of the population and the learning efficiency. Best solution or best fitness value can be obtained from the initial random threshold values, whose quality is evaluated by the correlation function. Experimental results have been examined on various numbers of thresholds. The results demonstrate that the proposed algorithm is more accurate and efficient than other four popular methods.**

*Index Terms*—**Cuckoo Search, Image segmentation, Multilevel thresholding, Metaheuristic**

## I. INTRODUCTION

Image segmentation plays a fundamental role in image understanding and computer vision, which partitioning a given image into several meaningful homogeneous regions. Among the last few decades, thresholding technique is one of the most common used segmentation method for various types of images, due to its simplicity, robustness and accuracy [1-9].

Basically, thresholding is used to identify and extract targets from the background on the basis of distribution of gray levels or texture in image objects[12]. If the object in an image is distinguished from the background by computing a single threshold value, it is termed as bi-level thresholding. Bi-level thresholding is easy to achieve using the conventional exhaustive methods. However, it performs unsatisfied on real-life images and remote sensing images [8]. As a result, multilevel thresholding is strongly required for such task. Multilevel thresholding is much more complex and usually needs some effective optimization methods. Nowadays, metaheuristic algorithms become popular for this problem [1-9, 11, 13-18], which take advantages of the convergence speed and accuracy when compared with the exhaustive method. The most common used algorithms include particle swarm optimization (PSO) [9, 14, 19], bacterial foraging optimization (BFO) [16, 21], differential evolution (DE) [18, 22], fire-fly algorithm (FA) [23], artificial bee colony (ABC) [1, 24], wind driven optimization (WDO) [12] and Cuckoo Search (CS) [11, 12].

Among all the involved algorithm, CS algorithm is outstanding due to its optimum performance in the multilevel thresholding framework [25]. Cuckoo is a population-based evolutionary algorithm which solves the structural optimization problems with the enhancement of Lévy flights. Various studies on multilevel thresholding based studies demonstrate that CS obviously outperforms other popular metaheuristic algorithms [11-13]. Authors in [26, 27] reported the best performance of CS when compared with genetic algorithm, PSO, ABC, DE, and BFO. Two comparative studies [11, 13] that use nature-inspired algorithms for multilevel thresholding show that CS offers the best performance when compared with PSO, DE, ABC and WDO. In addition, CS has also outperformed the recently emerging FA in solving multilevel thresholding problem in [28]. The comparative performance study in [11] has depicted the outstanding performance of CS using different objective functions as compared to other optimization algorithms with respect to the color image multilevel thresholding. Recently, an extensive study on CS algorithm using energy curve based different entropy objective criterions is discussed in [8].

In this paper, we propose a CS variant called Fully Informed Cuckoo Search (FICS) for the multilevel thresholding. The main contribution of this study is that we improved the CS through the fully informed neighborhood strategy. This strategy allows the individuals to learn from their neighbors, improving the learning efficiency and the cooperation of the population. The experimental results demonstrate that such modifications significantly improve the performance on multilevel

Xiaotao Huang, is with College of Electronic Science, National University of Defense Technology, Changsha, 410000, China. (email: xthuang@nudt.edu.cn)

Liang Shen, corresponding author, is with College of Electronic Science, National University of Defense Technology, Changsha, 410000, China. (email: shenliang16@nudt.edu.cn)

Chongyi Fan, is with College of Electronic Science, National University of Defense Technology, Changsha, 410000, China. (email: Chongyifan@sina.com)

Jiahua Zhu, is with College of Electronic Science, National University of Defense Technology, Changsha, 410000, China. (email: Zhujiahua1019@hotmail.com)

Sixian Chen, is with College of Electronic Science, National University of Defense Technology, Changsha, 410000, China. (email: 18696452470@163.com)

thresholding in terms of the objective function value, image quality measures on various level of thresholding, when compared with both the original CS and two latest modification of CS.

## II. MULTILEVEL THRESHOLDING

This section introduces the multilevel thresholding problem. Multilevel thresholding is a method that classifying the pixels of a given image into multiclass using a set of threshold values. The way to find the optimal thresholds is to maximize some discriminating criteria (also called objective functions). Given an image $I$ with gray levels ranging from $0$ to $L-1$, suppose there are $M$ thresholds $(t_1, t_2, ... t_M)$ which segment the gray levels of $I$ into $M+1$ classes: $C_0$ for $[0, t_1 - 1]$, $C_1$ for $[t_1, t_2 - 1]$, ..., $C_M$ for $[t_M, L-1]$, where $t_1 < t_2 < ... < t_M$. Then the definition to the problem of multilevel thresholding is given as follows.

$$(t_1^*, t_2^*, ... t_M^*) = \arg\max\left(f(t_1, t_2, ... t_M)\right) \quad (1)$$

where $f$ means the objective function. The most popular objective function is the between-class variance, which is based on Otsu's measure [29]. The Otsu's measure is known for its simplicity and affectivity with regard to uniformity and shape measures[9, 30, 31].

The Otsu's measure can be described as follows. With $t_0 = 0$ and $t_{M+1} = L$, the probability $P_i$ of gray level $i$ and the probabilities of class occurrence $\omega_m$ of a given image are firstly calculated as:

$$P_i = n_i / N \quad (2)$$

$$\omega_m = \sum_{i=t_m}^{t_{m+1}-1} P_i \quad (3)$$

where $n_i$ is the number of pixels of the $i$th level; $N = n_0 + n_1 + ... + n_{L-1}$ denotes the total number of pixels.

Then the objective function of Otsu's measure function is defined as:

$$f_1 = \sum_{m=0}^{M} \omega_m \left(\mu_m - \mu_T\right)^2 \quad (4)$$

where $\mu_m = \sum_{i=t_m}^{t_{m+1}-1} \frac{iP_i}{\omega_m}$ is the class mean levels; $\mu_T = \sum_{i=0}^{L-1} i \cdot P_i$ is the total mean level.

## III. CUCKOO SEARCH ALGORITHM

The cuckoo search (CS) algorithm is a population-based optimization algorithm that is originally proposed in [32]. It is developed according to two ideas: the cuckoo breeding behavior, and the Lévy flights. We first introduce the two ideas, and then give the implementation.

### A. Cuckoo Breeding Behavior

CS was inspired by the obligate brood parasitism of some cuckoo species initially. The female cuckoos of these species may lay their eggs in the nests of some host birds. Meanwhile, the horde bird may also recognize these eggs with a probability $p_a \in [0,1]$. Then, the horde bird throws the eggs from the nest or abandons the next to form a new nest.

To form the mathematical model, each individual of the host nests (with eggs) is assumed to be a candidate solution. Three main rules are concluded based on such behavior:

1). The number of available host nests (population size) is fixed, and the egg laid by a cuckoo is discovered by the host bird with a probability $p_a$.

2). Each Cuckoo lays one egg each time at a random host nest.

3). The nests with best-quality eggs will be kept to breed the next generations.

The first rule can be further illustrated that the nests are replaced by new nests with a probability $p_a$. Then, the cuckoo breeding behavior is modeled by (6) to generate new solutions:

$$v_j = \begin{cases} x_{i,j} + \varepsilon \cdot (x_{r1,j} - x_{r2,j}) & rand > p_a \\ x_{i,j} & Otherwise \end{cases} \quad (5)$$

where $x_i$, $i \in [1, 2, ... N]$ denotes the $i$th solution and $N$ is the population size; $j \in [1, D]$ is the $j$th dimension, where $D$ is the problem dimension (here, it corresponds to the number of thresholds $M$ in multilevel thresholding problem); $rand$ is a uniformly distributed random number. $v$ is the newly generated solution, which will replace $x_i$ only if its fitness is better than $x_i$.

### B. Lévy Flights

Lévy flights is associated with the flight behavior of some animals and insects. This behavior indicates that the distance and steps of these animals and insects jump or fly obey a Levy distribution. Such behavior is well adapted to generate new solution in the metaheuristic algorithms. In CS, Lévy flight is performed to generates random walks for the candidate solutions, where the step length is distributed according to the Lévy flight behavior, which can be formulated as (7):

$$v = x_i + \alpha L(\lambda) \cdot (x_i - best) \quad (6)$$

where $v$ is the newly generated solution, which will replace $x_i$ only if its fitness is better than $x_i$, $best$ is the current best solution; $\alpha$ is a scaling factor to control the step size; $L$ is the step size randomly drawn from the Lévy distribution. The Lévy distribution is given by:

$$L \sim \frac{\lambda \Gamma(\lambda) \sin(\pi \lambda / 2)}{\pi} \frac{1}{s^{1+\lambda}}$$

where $\lambda = 1.5$, $\Gamma(\lambda)$ stands for the standard gamma function, and the distribution is valid for large steps $s > 0$ according to [33].

### C. Implementation

In the beginning of a run, all individuals are generated randomly within the boundaries of the parameters. All the individuals (candidate solutions) undergo an iterative process where their positions are updated according to Lévy flights and breeding behavior successively after the initialization step. The best solution obtained at the last generation provides the optimal solution. The pseudo code of CS is shown in Algorithm 1.

## IV. THE PROPOSED ALGORITHM

Although CS shows remarkable performance in multilevel thresholding, recent studies point that the original CS may also offer low searching capacity [34] and its convergence rate can also be improved in multilevel thresholding [20]. Next, we introduced the idea of neighborhood strategy to CS and proposed an improved CS using a fully informed neighborhood strategy.

For the individuals in the cuckoo breeding behavior of the original CS as (5) shows, each new solution is generated around the position of the individual itself. As a population-based algorithm, the self-mutation strategy lacks of cooperation and interaction between the individuals, which may result in low learning efficiency. The neighborhood strategy is designed for such a problem. It is known that the neighborhood strategy can effectively improve the performance of the swarm-based algorithms [35-38]. Therefore, we introduce a neighborhood strategy called the fully informed strategy to improve the intraspecific cooperation and interaction of the population in CS.

The fully informed strategy is firstly proposed to improve PSO in [10]. It aims to define a better solution for each individual to learn from (by using the good experience of its neighbors). Supposing there are $Ne = 2*n+1$ neighbors for the $i$th individual $x_i$, which can be denoted as $x_{i-n}$, $x_{i-n+1}$, ..., $x_{i+n}$ [1]. Then, the cuckoo breeding behavior refers to (5) is modified as:

$$v_j = \begin{cases} FI + \varepsilon \cdot (x_{r1,j} - x_{r2,j}) & rand > p_a \\ x_{i,j} & Otherwise \end{cases} \quad (7)$$

where $FI$ denotes the information learned from the neighbors, which is defined as (9)

$$FI = s \cdot \sum_{k=i-n}^{i+n} w_k \cdot x_{k,j} \quad (8)$$

here, $w_k$ is the weights of the $k$th neighbor calculated as (10) for a maximization problem. It determines the weight of the influence from the corresponding neighbor. $s$ is the scaling factor, or a normalization operator, which is calculated by (11).

ALGORITHM I. PSEUDO CODE OF CS

| | |
|---|---|
| 1 | **Begin** |
| 2 | Objective function $f(x), x = (x_1, x_2, ... x_M)$ ; |
| 3 | Initialize a population of $Np$ host nests $x_i, (i = 1, 2...Np)$ |
| | Update the best solution $best$ , initialize the counter $Cnt = 0$ ; |
| 4 | **While** $Cnt < MaxFEs$ |
| 5 | \** *Lévy flights* **\ |
| 6 | **For** $i = 1$ to $Np$ **do** |
| 7 | Draw a step vector $Levy$ which obeys a Lévy distribution |
| 8 | Generate a new solution: $v = x_i + \alpha \oplus Levy \cdot (x_i - best)$ |
| 9 | Evaluate the new solution $v$ using the objective function; |
| 10 | **If** $v$ is better than $x_i$ |
| 11 | $x_i = v$ ; |
| 12 | **End if** |
| 13 | **End for** |
| 14 | \** *Cuckoo breeding behavior* **\ |
| 15 | **For** $i = 1$ to $D$ |
| 16 | **For** $j = 1$ to $D$ |
| 17 | **If** $rand < p_a$ **Then** |
| 18 | $v_j = x_{i,j} + rand \cdot (x_{r1,j} - x_{r2,j})$ |
| 19 | **End if** |
| 20 | **End for** |
| 21 | Evaluate the new solution $v$ using the objective function; |
| 22 | **If** $v$ is better than $x_i$ |
| 23 | $x_i = v$ ; |
| 24 | **End if** |
| 25 | **End for** |
| 26 | Update the global best solution $best$ |
| 27 | **End for** |
| 28 | **End While** |
| 29 | **End** |

$$w_k = rand \cdot f_k \quad (9)$$

$$s = 1 / \sum_{k=i-n}^{i+n} w_k \quad (10)$$

In the strategy, the new solution is generated based on $FI$ rather than the solution itself $x_i$. In other words, the modified breeding behavior can generate new solution (nest) according to both the solution itself and the information from the neighbors.

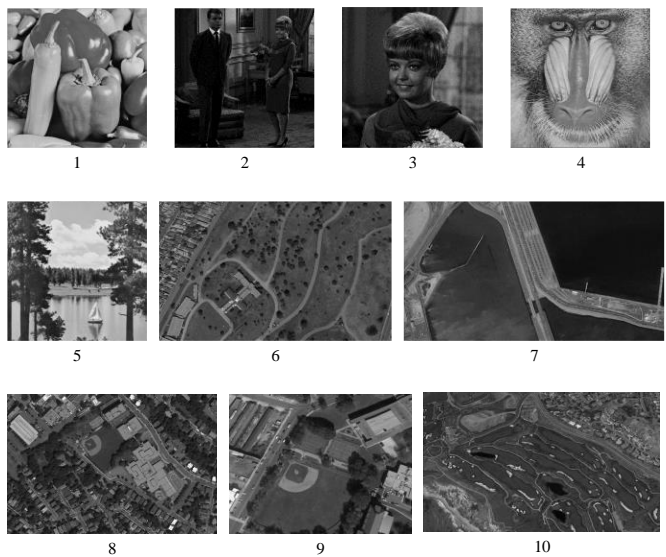

1　　2　　3　　4

5　　6　　7

8　　9　　10

Fig. 1.  Images used in the experiments.

**Algorithm II.  Modified Cuckoo Breeding Behavior**

| | |
|---|---|
| 1 | \** *Cuckoo breeding behavior* **\ |
| 2 | **Begin** |
| 3 | **For** $i = 1$ to $Np$ **do** |
| 4 | **For** $j = 1$ to $D$ |
| 5 | **If** $rand < p_a$ **Then** |
| 6 | Generate the $FI$ for the $j$th dimension using $$FI = s \cdot \sum_{k=i-n}^{i+n} w_k \cdot x_{k,j}$$ |
| 7 | Generate the $j$th dimension for the new solution using $v_j = FI + rand \cdot (x_{r1,j} - x_{r2,j})$ |
| 8 | **End if** |
| 9 | **End for** |
| 10 | Evaluate the new solution $v$ using the objective function; |
| 11 | **If** $v$ is better than $x_i$ |
| 12 | $x_i = v$ ; |
| 13 | **End if** |
| 14 | **End for** |
| 15 | Update the global best solution $best$ |
| 16 | **End for** |
| 17 | **End** |

In such a case, the new solution is more likely to be high-quality, and the exploitation around such a solution will promote the population to search larger potential space.

Due to the capability of dispersing the individuals towards a random position (which enhances the exploration), this strategy makes significant sense for the algorithm in terms of escaping from the local optimal and avoiding the premature convergence. The implementation of the proposed modified cuckoo breeding behavior is given in Algorithm 2, where the modifications are marked with red number.

## V. EXPERIMENTAL SETUP

This section evaluates the performance of the proposed algorithm, whose advantage is illustrated by comparing with the following four algorithms: FIPSO, CS, CS-MA and CS-MC, where CS is the standard CS, FIPSO and CS-MA are the two basic studies of our method, and CS-MC is a recently published improved version of CS. To the best of our knowledge, FIPSO has never been applied for multilevel thresholding before, here we did this work in this paper and compared it with our algorithm. CS, CS-MA and CS-MC have been well applied for multilevel thresholding which have been demonstrated the best performance in the corresponding references in TABLE I.

The population size was set as 30 for all algorithms; other particular parameters of the involved algorithms were set as TABLE I shows (according to the corresponding references). 30 independent runs were carried out for each image of the algorithms on each number of thresholds in order to reduce random errors. The results of the mean objective value and the standard deviation are recorded for comparison. [2]The number of thresholds $M$ is set to be 3, 7, 11 and 15. All methods are carried out with the same calculation limitation and the maximum number of function evaluations was set as $1200*M$ (it corresponds to 120, 280, 440 and 600 maximum iterations for $M$ = 3, 7, 11 and 15 respectively). The total gray scale level $L = 255$ and the values of the solution are directly transformed into the nearest integer in the optimization process. The tested images are shown in Fig. 1, where the first five images (images 1-5) are popular real-life test images in image process, and the other five images are optical remote sensing images.

TABLE I
PARAMETERS AND REFERENCES OF THE COMPARED ALGORITHMS

| Algorithm | Parameters | Value | Reference |
|---|---|---|---|
| **FIPSO** | Cognitive, social acceleration | 2, 2 | [10] |
| | Inertial weight | 0.95-0.4 | |
| **CS** | Mutation probability value | 0.25 | [11] |
| | Scale factor | 1.5 | |
| | Step size | 1 | |
| **CS-MA** | Mutation probability value | 0.25 | [12] |
| | Scale factor | 1.5 | |
| **CS-MC** | Mutation probability value | 0.5 | [20] |
| | Scale factor | 1.5 | |
| **FICS** | Mutation probability value | 0.5 | This paper |
| | Scale factor | 1.5 | |
| | Neighbors $Ne$ | 3 | |

## VI. RESULTS

This section evaluates the performance of the proposed algorithm. On the one hand, the involved algorithms are compared according to the objective function values. On the other hand, we also verified the performance using two popular quality measures. The nonparametric test is used for the comparison for rigorousness.

### A. Comparison on Objective Function Values

The objective function value obtained by the involved algorithms directly shows the algorithm's performance, since the multilevel thresholding is to maximize the given objective function. TABLE II show the results, where "Std" means the standard deviation and "h" means the pairwise comparison results between the corresponding algorithm and the proposed one. The best mean value is shown in bold.

As is illustrated, the proposed algorithm obtained the best mean value in 39 blocks and 40 blocks (both 40 in total) in TABLE II, which represented the best overall performance.

To verify whether the results generated by FICS are significantly different from the compared algorithms, we performed the nonparametric statistical Wilcoxon rank sum test [39] to perform rigorous comparisons between FICS and its peers. The test was conducted at 5% significance level. The value of h indicates whether the performance of FICS is better (i.e., h = "+"), insignificant (i.e., h = "="), or worse ((i.e., h = "-")) than the compared algorithm at the statistical level. The Comparison results are summed up in the button of the tables, where W/T/L means the total times FICS wins/ties/loses the comparison when compared with the corresponding algorithm.

We can observe from all the two tables that FICS wins most of the blocks in the comparison with any other algorithms, and only loses to FIPSO in one block (3-level thresholding on image 2). The most competitive algorithm to FICS is CS-MA on both the two measures. Compared with CS-MA, FICS shows close performance in 18 blocks and wins all other 22 blocks.

The Friedman test is further employed to rank all the algorithms [39, 40] and evaluate the difference of their performance. The Friedman test captured the result of locating optimal thresholds on all ten images used in the experiments according to all observed numbers of thresholds [31]. It allows us to highlight those ones whose performances are statistically different which offers valid procedures to rank the involved algorithms. The ranking is performed on each level of thresholding. Therefore, four test were conducted and 10 ($n = 10$) variables were used in each comparison in each test, where $n = 10$ denotes the number of images, the significance level is considered to be 0.05.

Fig. 2 to Fig. 5 show the ranking results of the four level thresholding respectively, where the center circle denotes the average ranks and the lines indicate the confidence intervals. Higher rank values represent a better performance and two algorithms are regarded to be significantly different if there is no overlap between any intervals of the algorithms. Generally speaking, all algorithms shows close rankings in the corresponding level thresholding, where the proposed algorithm ranks the best on all comparisons. The most competitive algorithms to FICS are FIPSO, CS-MC, CS-MA and CS-MA on 3, 7, 11 and 15 level thresholding, respectively. However, all these algorithms are not comparative in other cases.

### B. QUALITY MEASURES OF SEGMENTED IMAGE

To verify the performance of FICS, we introduce two quantitative comparison measures: Peak Signal to Noise Ratio (PSNR) and Structural SIMilarity (SSIM), to evaluate the corresponding image segmentation performance of the results obtained in subsection A. PSNR and SSIM are the most popular quality measures in multilevel thresholding [13, 14, 31]. The PSNR of segmented image indicates the accuracy of the reconstructed image calculated by (12) and (13), and SSIM shows the visual similarity of the reconstructed image against the original image according to the degradation of structural information given by (14).

$$PSNR = 10\log_{10}\left(\frac{255^2}{MSE}\right) \qquad (11)$$

$$MSE = \frac{1}{MN}\sum_{i=1}^{M}\sum_{j=1}^{N}\left[x(i,j) - y(i,j)\right]^2 \qquad (12)$$

$$SSIM(x,y) = \frac{(2\mu_x\mu_y + c_1)(2\sigma_{xy} + c_2)}{(\mu_x^2 + \mu_y^2 + c_1)(\sigma_x^2 + \sigma_y^2 + c_2)} \qquad (13)$$

where $x$ is the original image and $y$ is the reconstructed image, $u_x$ and $u_y$ are the mean intensities of $x$ and $y$; $\sigma_x^2$ and $\sigma_y^2$ stand for the variance of $x$ and $y$; $\sigma_{xy}$ is the covariance of $x$ and $y$. $c_1 = 6.5025$, $c_2 = 58.5225$ [31].

Without loss of generality, two practical images: Image 1 and Image 2, and two remote sensing images: Image 6 and Image 7 were chosen to illustrate the overall performance. The results of the above two parameters are listed in TABLE IV.

It can be observed that FICS won 11 times in total 16 comparisons on both PSNR and SSIM, which obviously outperforms other algorithms.

## VII. CONCLUSION

This paper proposed a fully informed cuckoo search algorithm for multilevel thresholding image segmentation. The main contribution is a successively introduced full informed strategy, which improved the performance of cuckoo search on different level of thresholding for various kinds of images. The experiments demonstrated the effectiveness of our work when comparing with four other popular metaheuristic algorithms in terms of the mean objective function value. Strict rankings based on the Friedman test clearly demonstrated the superiority of the proposed algorithm on all level thresholding. In addition, the performace of proposed thresholding technique is also validated on image quality measures such as PSNR and SSIM.

APPENDIX

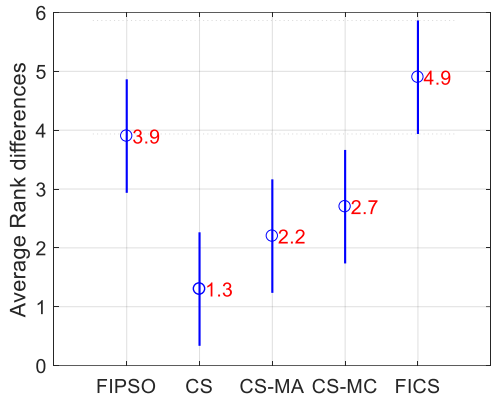

Fig. 2. Ranking results of the Friedman test of 3-level thresholding.

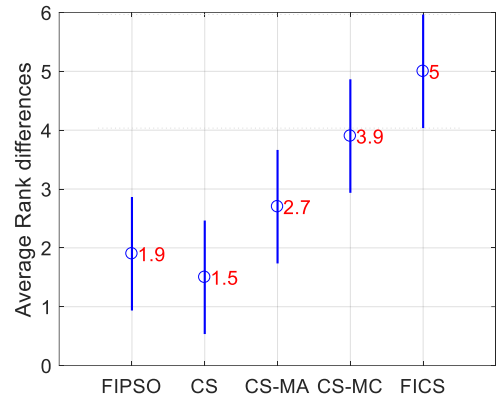

Fig. 3. Ranking results of the Friedman test of 7-level thresholding.

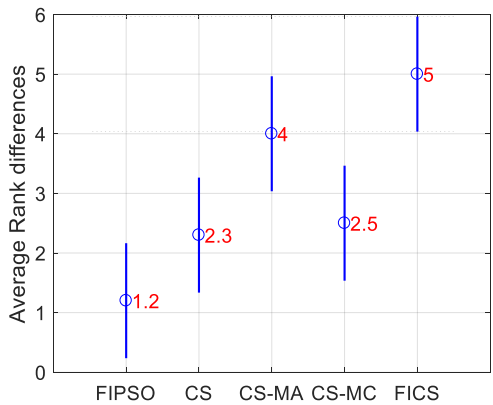

Fig. 4. Ranking results of the Friedman test of 11-level thresholding.

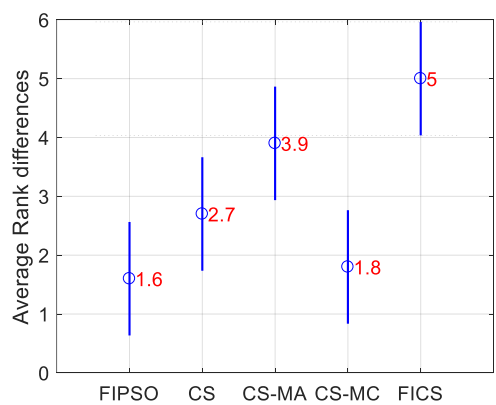

Fig. 5. Ranking results of the Friedman test of 15-level thresholding.

TABLE II
COMPARISON OF THE MEAN OBJECTIVE FUNCTION VALUES ON OTSU'S MEASURE

| Image | $M$ | FIPSO | | | CS | | | CS-MA | | | CS-MC | | | FICS | (control) |
|---|---|---|---|---|---|---|---|---|---|---|---|---|---|---|---|
| | | Mean | (Std) | h | Mean | (Std) | h | Mean | (Std) | h | Mean | (Std) | h | Mean | (Std) |
| Image 1 | 3 | 2703.564 | (2.47E-02) | + | 2703.535 | (9.77E-02) | + | 2703.530 | (9.36E-02) | = | 2703.552 | (3.74E-02) | = | **2703.572** | (0.00E+00) |
| | 7 | 2846.650 | (1.04E+00) | + | 2848.005 | (5.02E-01) | + | 2848.229 | (4.05E-01) | + | 2848.581 | (2.09E-01) | + | **2848.863** | (4.77E-02) |
| | 11 | 2874.661 | (1.24E+00) | + | 2877.418 | (7.74E-01) | = | 2877.776 | (5.45E-01) | + | 2876.651 | (9.98E-01) | + | **2878.801** | (8.05E-02) |
| | 15 | 2885.009 | (4.92E-01) | + | 2887.547 | (4.83E-01) | + | 2888.077 | (4.47E-01) | + | 2886.687 | (8.16E-01) | + | **2888.953** | (1.25E-01) |
| Image 2 | 3 | **904.6902** | (1.66E-02) | − | 904.2343 | (5.38E-01) | = | 904.3875 | (3.14E-01) | = | 904.0448 | (9.35E-01) | = | 904.6897 | (1.55E-02) |
| | 7 | 975.4630 | (5.24E-01) | = | 973.3014 | (1.48E+00) | = | 973.8184 | (1.59E+00) | + | 975.1167 | (1.08E+00) | + | **976.4183** | (4.53E-02) |
| | 11 | 986.4797 | (4.46E-01) | = | 985.6981 | (8.79E-01) | + | 986.7955 | (7.50E-01) | = | 986.6171 | (6.25E-01) | + | **988.1642** | (9.74E-02) |
| | 15 | 991.5273 | (3.32E-01) | = | 991.1284 | (6.22E-01) | + | 991.4965 | (5.16E-01) | = | 991.3661 | (6.02E-01) | + | **993.0640** | (2.81E-01) |
| Image 3 | 3 | 1444.736 | (8.43E-03) | + | 1444.506 | (2.98E-01) | + | 1444.623 | (1.89E-01) | = | 1444.645 | (1.27E-01) | = | **1444.741** | (2.31E-13) |
| | 7 | 1539.941 | (8.79E-01) | + | 1539.205 | (1.41E+00) | = | 1539.964 | (1.19E+00) | + | 1540.578 | (1.02E+00) | + | **1541.842** | (3.35E-02) |
| | 11 | 1558.567 | (7.22E-01) | = | 1558.585 | (1.13E+00) | + | 1559.599 | (7.75E-01) | + | 1558.806 | (8.03E-01) | = | **1561.021** | (1.82E-01) |
| | 15 | 1565.588 | (4.95E-01) | = | 1566.158 | (7.12E-01) | + | 1566.871 | (5.76E-01) | + | 1565.779 | (7.40E-01) | = | **1568.295** | (2.63E-01) |
| Image 4 | 3 | 1639.511 | (5.70E-02) | = | 1639.419 | (1.17E-01) | + | 1639.445 | (9.92E-02) | + | 1639.523 | (2.66E-02) | + | **1639.532** | (2.33E-03) |
| | 7 | 1746.314 | (7.74E-01) | + | 1746.883 | (6.96E-01) | + | 1747.186 | (5.73E-01) | + | 1747.785 | (2.55E-01) | + | **1748.259** | (3.63E-02) |
| | 11 | 1767.353 | (6.55E-01) | + | 1769.202 | (8.14E-01) | + | 1769.740 | (5.23E-01) | = | 1768.790 | (9.64E-01) | + | **1770.810** | (7.88E-02) |
| | 15 | 1775.582 | (5.62E-01) | + | 1777.346 | (7.19E-01) | + | 1777.944 | (4.94E-01) | + | 1776.540 | (7.71E-01) | + | **1779.035** | (6.12E-02) |
| Image 5 | 3 | 4113.762 | (1.49E-02) | = | 4113.713 | (1.09E-01) | + | 4113.737 | (5.48E-02) | + | 4113.766 | (5.05E-03) | + | **4113.767** | (0.00E+00) |
| | 7 | 4249.444 | (9.85E-01) | + | 4250.408 | (6.82E-01) | + | 4250.661 | (2.64E-01) | + | 4250.850 | (2.27E-01) | + | **4251.112** | (5.49E-02) |
| | 11 | 4274.505 | (1.22E+00) | + | 4276.665 | (6.05E-01) | = | 4277.318 | (5.30E-01) | = | 4276.591 | (1.03E+00) | + | **4278.195** | (5.80E-02) |
| | 15 | 4283.818 | (6.84E-01) | + | 4286.280 | (5.79E-01) | + | 4286.863 | (3.73E-01) | + | 4285.374 | (7.98E-01) | + | **4287.685** | (1.16E-01) |

| | | | | | | | | | | | |
|---|---|---|---|---|---|---|---|---|---|---|---|
| Image 6 | 3 | 554.5900 (3.12E-02) | = | 554.4333 (1.74E-01) | = | 554.4675 (1.21E-01) | = | 554.5568 (6.45E-02) | = | **554.6008** (4.62E-13) |
| | 7 | 616.5303 (5.93E-01) | + | 615.5820 (1.13E+00) | + | 616.2803 (8.84E-01) | + | 617.0360 (6.44E-01) | + | **617.8526** (6.28E-02) |
| | 11 | 630.2058 (5.94E-01) | + | 630.3401 (8.14E-01) | = | 630.9963 (5.68E-01) | = | 630.5737 (7.79E-01) | = | **632.3313** (1.31E-01) |
| | 15 | 635.7244 (4.35E-01) | + | 635.7565 (4.63E-01) | + | 636.4803 (4.36E-01) | + | 635.5902 (6.20E-01) | + | **637.6898** (1.74E-01) |
| Image 7 | 3 | 1513.281 (1.97E-02) | + | 1513.036 (3.49E-01) | = | 1513.193 (9.65E-02) | = | 1513.174 (1.19E-01) | + | **1513.287** (6.93E-13) |
| | 7 | 1583.173 (6.02E-01) | + | 1581.830 (1.45E+00) | + | 1582.477 (1.10E+00) | + | 1583.505 (6.67E-01) | + | **1584.462** (5.70E-02) |
| | 11 | 1594.583 (4.15E-01) | + | 1594.528 (6.98E-01) | + | 1595.110 (5.60E-01) | = | 1594.866 (5.26E-01) | + | **1596.580** (2.20E-01) |
| | 15 | 1600.064 (4.25E-01) | + | 1600.029 (6.87E-01) | + | 1600.835 (5.41E-01) | + | 1599.883 (7.61E-01) | + | **1602.144** (2.33E-01) |
| Image 8 | 3 | 1403.174 (3.15E-01) | + | 1403.119 (2.60E-01) | = | 1403.148 (3.56E-01) | + | 1403.087 (3.56E-01) | + | **1403.358** (1.48E-01) |
| | 7 | 1542.970 (1.17E+00) | + | 1543.810 (9.90E-01) | + | 1544.180 (6.23E-01) | + | 1544.484 (5.29E-01) | + | **1545.094** (5.23E-02) |
| | 11 | 1565.415 (7.86E-01) | = | 1566.894 (7.53E-01) | + | 1567.567 (5.79E-01) | + | 1566.865 (1.07E+00) | + | **1568.876** (1.66E-01) |
| | 15 | 1574.157 (6.38E-01) | = | 1575.905 (4.00E-01) | + | 1576.223 (3.61E-01) | + | 1575.008 (6.58E-01) | + | **1577.273** (1.59E-01) |
| Image 9 | 3 | 1283.569 (9.98E-03) | + | 1283.457 (2.94E-01) | = | 1283.485 (1.04E-01) | + | 1283.523 (8.85E-02) | = | **1283.574** (0.00E+00) |
| | 7 | 1384.286 (1.29E+00) | + | 1384.609 (8.53E-01) | = | 1385.241 (7.37E-01) | + | 1385.501 (7.24E-01) | + | **1386.164** (3.25E-02) |
| | 11 | 1404.916 (1.04E+00) | + | 1406.808 (8.54E-01) | + | 1407.250 (6.59E-01) | + | 1406.705 (7.67E-01) | + | **1408.521** (1.38E-01) |
| | 15 | 1413.563 (4.91E-01) | = | 1415.037 (4.98E-01) | + | 1415.512 (3.95E-01) | = | 1414.403 (5.79E-01) | + | **1416.555** (1.27E-01) |
| Image 10 | 3 | 808.0678 (2.52E-02) | + | 807.9813 (1.32E-01) | = | 807.9915 (9.36E-02) | = | 808.0082 (7.15E-02) | + | **808.0769** (1.81E-03) |
| | 7 | 884.8145 (7.86E-01) | + | 884.6663 (1.00E+00) | = | 885.2203 (6.48E-01) | + | 885.3498 (7.66E-01) | + | **886.5004** (7.40E-02) |
| | 11 | 901.4185 (6.10E-01) | + | 901.9566 (9.07E-01) | = | 902.7234 (6.20E-01) | + | 902.5248 (8.03E-01) | + | **904.0290** (1.07E-01) |
| | 15 | 907.9853 (5.07E-01) | = | 908.2501 (6.00E-01) | + | 908.9904 (5.62E-01) | = | 908.0400 (7.09E-01) | + | **910.2729** (1.11E-01) |
| **Total** | **W/T/L** | **28/11/1** | | **26/14/0** | | **22/18/0** | | **29/11/0** | | **/** |

TABLE IV
COMPARISON OF THE MEAN PSNR AND SSIM

| | | PSNR Comparison on Otsu's measure | | | | | PSNR Comparison on Kapur's measure | | | | |
|---|---|---|---|---|---|---|---|---|---|---|---|
| Image | K | FIPSO | CS | CS-MA | CS-MC | FICS | FIPSO | CS | CS-MA | CS-MC | FICS |
| Im1 | 3 | 25.1295 | 25.1280 | 25.1277 | **25.1739** | 25.1304 | 25.0523 | 25.0481 | 25.0448 | **25.1141** | 25.0540 |
| | 7 | 30.5981 | 30.7061 | 30.7316 | **30.7986** | 30.7808 | 29.9804 | 30.0826 | 30.1161 | 30.1432 | **30.1542** |
| | 11 | 33.5753 | 34.0265 | 34.0656 | 33.9470 | **34.2564** | 33.1895 | 33.7291 | 33.8282 | 33.7361 | **34.0004** |
| | 15 | 35.5211 | 36.1780 | 36.3449 | 35.9962 | **36.6068** | 34.8523 | 35.7903 | 35.8288 | 35.7462 | **36.3012** |
| Im2 | 3 | 28.3293 | 28.3084 | 28.3132 | **28.4160** | 28.3304 | **22.9808** | 22.9645 | 22.9627 | 22.9759 | 22.9702 |
| | 7 | 34.1908 | 33.8105 | 33.9021 | 34.1983 | **34.3649** | **29.4617** | 28.9275 | 28.8102 | 28.8002 | 28.7106 |
| | 11 | 36.7597 | 36.4952 | 36.8457 | 36.8343 | **37.3149** | 32.7752 | **33.0234** | 32.9069 | 32.8419 | 32.8782 |
| | 15 | 38.6578 | 38.5207 | 38.7434 | 38.7221 | **39.5917** | 34.4409 | 35.9424 | 36.1211 | 34.8884 | **36.1878** |
| Im6 | 3 | 28.5470 | 28.5342 | 28.5399 | **28.6742** | 28.5470 | 25.6679 | 25.6659 | **25.6957** | 25.6925 | 25.6867 |
| | 7 | 33.5108 | 33.3684 | 33.4734 | 33.6441 | **33.7112** | 29.3445 | **29.5263** | 29.3757 | 28.8932 | 28.8543 |
| | 11 | 36.2709 | 36.3196 | 36.5107 | 36.4689 | **36.9296** | 31.9781 | **32.5507** | 32.0732 | 32.1251 | 32.2946 |
| | 15 | 38.2163 | 38.2111 | 38.5614 | 38.2154 | **39.1897** | 34.0427 | 34.8356 | 35.0773 | 34.7380 | **35.0885** |
| Im7 | 3 | 28.2665 | 28.2569 | 28.2637 | **28.3411** | 28.2664 | 24.6810 | 24.6842 | 24.6874 | **24.6937** | 24.6758 |
| | 7 | 33.8147 | 33.6045 | 33.7100 | 33.9433 | **34.0328** | 29.4420 | 29.5558 | 29.8372 | 29.9860 | **30.0223** |
| | 11 | 36.2314 | 36.1926 | 36.3591 | 36.3427 | **36.8135** | 32.9942 | 33.7411 | 34.1647 | 33.4829 | **34.4926** |
| | 15 | 38.0903 | 38.0809 | 38.4494 | 38.0676 | **39.1146** | 34.3803 | 35.3188 | 35.3016 | 34.9791 | **35.4305** |
| | | 0 | 0 | 0 | 5 | **11** | 2 | 3 | 1 | 2 | **8** |

| | | SSIM Comparison on Otsu's measure | | | | | SSIM Comparison on Kapur's measure | | | | |
|---|---|---|---|---|---|---|---|---|---|---|---|
| Image | $K$ | FIPSO | CS | CS-MA | CS-MC | FICS | FIPSO | CS | CS-MA | CS-MC | FICS |
| Im1 | 3 | 0.812944 | 0.812918 | 0.812837 | **0.813890** | 0.813023 | 0.811158 | 0.811243 | 0.811174 | **0.812521** | 0.811210 |
| | 7 | 0.919456 | 0.920239 | 0.920788 | **0.921098** | 0.920179 | 0.908753 | 0.910209 | **0.910427** | 0.910416 | 0.910277 |
| | 11 | 0.955840 | 0.959896 | 0.960243 | 0.959738 | **0.962104** | 0.951434 | 0.956500 | 0.957357 | 0.956341 | **0.958810** |
| | 15 | 0.971214 | 0.975424 | 0.976313 | 0.974488 | **0.978032** | 0.965884 | 0.972208 | 0.972391 | 0.972176 | **0.975454** |
| Im2 | 3 | 0.771208 | 0.772690 | 0.771862 | **0.774179** | 0.770761 | **0.549875** | 0.549112 | 0.549320 | 0.549654 | 0.549560 |
| | 7 | 0.897624 | 0.889388 | 0.891525 | 0.895095 | **0.898654** | **0.741089** | 0.722726 | 0.718725 | 0.718057 | 0.715328 |
| | 11 | 0.929041 | 0.926192 | 0.930153 | 0.925741 | **0.936471** | 0.827559 | **0.831293** | 0.827749 | 0.827634 | 0.825095 |
| | 15 | 0.952162 | 0.951789 | 0.954407 | 0.950210 | **0.960939** | 0.865840 | 0.893417 | 0.895156 | 0.871654 | **0.895665** |
| Im6 | 3 | 0.900230 | 0.900878 | 0.900782 | **0.902429** | 0.898849 | 0.768557 | 0.768207 | **0.769717** | 0.769447 | 0.769399 |
| | 7 | 0.969828 | 0.967750 | 0.968837 | 0.971193 | **0.971827** | 0.877602 | **0.884233** | 0.878368 | 0.868127 | 0.866430 |
| | 11 | 0.984176 | 0.984251 | 0.985002 | 0.985034 | **0.987307** | 0.931328 | **0.939887** | 0.930859 | 0.932902 | 0.934502 |
| | 15 | 0.990334 | 0.990144 | 0.990982 | 0.990102 | **0.992641** | 0.958171 | 0.966265 | 0.968663 | 0.965761 | **0.969534** |
| Im7 | 3 | 0.874624 | 0.874184 | 0.873891 | **0.875225** | 0.874684 | 0.822707 | 0.822631 | 0.822826 | **0.823348** | 0.822868 |
| | 7 | 0.944874 | 0.941204 | 0.943065 | 0.945263 | **0.946385** | 0.889366 | 0.890439 | 0.893794 | 0.894992 | **0.895017** |
| | 11 | 0.812944 | 0.812918 | 0.812837 | **0.813890** | 0.813023 | 0.928672 | 0.935705 | 0.939051 | 0.933757 | **0.941344** |
| | 15 | 0.919456 | 0.920239 | 0.920788 | **0.921098** | 0.920179 | 0.943997 | 0.948407 | **0.948714** | 0.946403 | 0.948406 |
| | | 0 | 0 | 0 | 5 | **11** | 2 | 3 | 3 | 2 | **6** |

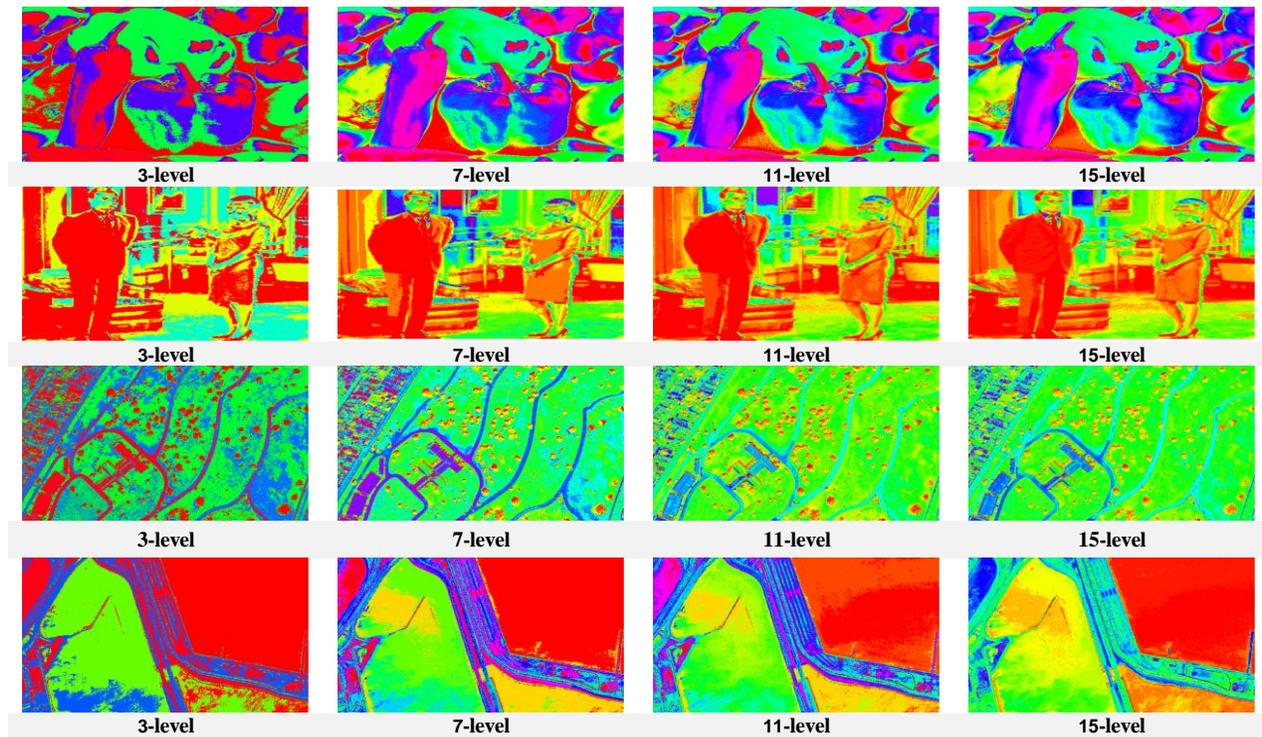

Fig. 5. Segmentation results using the proposed algorithm on Otsu's measure.